# Usability of a Robot's Realistic Facial Expressions and Peripherals in Autistic Children's Therapy

Jamy Li
*University of Twente*
Enschede, The Netherlands
j.j.li@utwente.nl

Daniel Davison
*University of Twente*
Enschede, The Netherlands
d.p.davison@utwente.nl

Bob Schadenberg
*University of Twente*
Enschede, The Netherlands
b.r.schadenberg@utwente.nl

Pauline Chevalier
*Istituto Italiano di Tecnologia*
Genoa, Italy
pauline.chevalier@iit.it

Alyssa Alcorn
*University College London*
London, UK
a.alcorn@ucl.ac.uk

Alria Williams
*University College London*
London, UK
alria.williams@ucl.ac.uk

Suncica Petrovic
*Serbian Society of Autism*
Belgrade, Serbia
suncica.petrovic@yahoo.com

Snezana Babovic Dimitrijevic
*Serbian Society of Autism*
Belgrade, Serbia
babovicnena@gmail.com

Jie Shen
*Imperial College London*
London, UK
jie.shen07@imperial.ac.uk

Liz Pellicano
*Macquerie University*
Sydney, Australia
liz.pellicano@mq.edu.au

Vanessa Evers
*University of Twente*
Enschede, The Netherlands
v.evers@utwente.nl

*Abstract*—Robot-assisted therapy is an emerging form of therapy for autistic children, although designing effective robot behaviors is a challenge for effective implementation of such therapy. A series of usability tests assessed trends in the effectiveness of modelling a robot's facial expressions on realistic facial expressions and of adding peripherals enabling child-led control of emotion learning activities with autistic children. Nineteen autistic children interacted with a small humanoid robot and an adult therapist in several emotion-learning activities that featured realistic facial expressions modelled on either a pre-existing database or live facial mirroring, and that used peripherals (tablets or tangible 'squishies') to enable child-led activities. Both types of realistic facial expressions by the robot were less effective than exaggerated expressions, with the mirroring being unintuitive for children. The tablet was usable but required more feedback and lower latency, while the tactile tangibles were engaging aids.

*Keywords—human-robot interaction; autism spectrum disorders; early childhood development; behavior design; tablets; tangible user interfaces; facial expressions*

## I. INTRODUCTION

A growing literature advocates the use of robots in therapeutic settings for autistic children [1]. Researchers have explored a variety of different scenarios, roles and appearances for a robot in robot-assisted autism therapy. Yet few guidelines exist for how interactions with robots should be designed. This issue is particularly important given that special consideration is needed for designing environments and technologies for autistic people [2,3]. Are robot facial expressions effective when they are designed to be realistic like how a person smiles rather than "caricatured" (i.e., exaggerated) like how a cartoon character smiles? Can a tablet be used with autistic children to support interactivity and if so, how should its interface be designed? How do children react to tangible "squishies" while interacting with a robot? That is, are robot facial expressions that have ecologically validity and tablet or tangible peripherals suitable for use in robot-assisted therapy with autistic children?

The present research addressed these questions via a series of usability studies evaluating the response of autistic children to a robot's realistic facial expressions and touchable peripherals. Our goal was to determine the validity of realistic facial expressions on a robot that are more ecologically valid and peripherals that provide more control to autistic children, which is thought to be important in the design of robot-assisted interventions [4]. Past literature in emotional facial expression found autistic adults believe exaggerated emotions are more realistic than average emotions [5], but this comparison has not been assessed with autistic children. Previous work explores the benefits of tangible interfaces [6,7] and tablets [8] for autistic children, but few studies have assessed them in conjunction with a robot. Evaluating realism of robot behavior and child-initiated interaction through a tablet or tangible device can aid roboticists in creating effective robot systems for autism therapy.

The methodologies of usability studies with autistic children differ. Khan et al. [9] tested the usability of two smartphone apps with 50 autistic young adults or autistic children and their guardians, who filled out quantitative and open-ended questions about ease of use, learnability, feedback, help documentation and appeal. The most problematic issue for both apps was lack of feedback. Weiss et al. [10] tested the usability of collaborative gaming on a multi-touch table with four pairs of autistic boys aged 9-13, who filled out a survey and were interviewed afterwards. The collaborative patterns in the games effectively engaged children and appeared to promote more ecologically valid playing than playing alone. These usability studies evaluate new technology prototypes for autistic children using usability concepts like feedback and learnability (cf. [11]), which we employ in the current work.

## II. MATERIALS AND METHODS

*Participants.* Nineteen autistic children (15 M, 4 F) between 4 and 16 years old participated in usability tests in May-June 2018. The study was held in two locations: 8 autistic children (2 female, mean age of 8.5 years) of which 7 had previously interacted with the robot or therapist during a prior study were from a London school for special education; 11 autistic children (2 female, mean age of 8 years and 8 months) of which 10 had interacted with the robot or therapist during a prior study were from the Serbia Society of Autism community in Belgrade. Sessions took 15-25 minutes and children completed 1, 2 or 3 sessions over the course of the two-week study. A variable number of sessions were held until the child mastered the activities. Five participants engaged in prior sessions interacting with the robot.

*Materials.* The Zeno humanoid robot was set on a table in a small room in a quiet area in the school. The child sat opposite to the robot and the adult sat on the side of the table between the child and robot. The child's tablet was enclosed in a protective case and was placed on the table in front of the child. A camera recorded an overview of the interaction showing the child, robot and adult. In London only, two additional cameras recorded an over-the-shoulder view of the robot and a close-up of the child's face. The setup is illustrated in Figure 1. A researcher hidden behind a room divider took notes. Theme-based analysis of results was done using interview notes and viewing of videos.

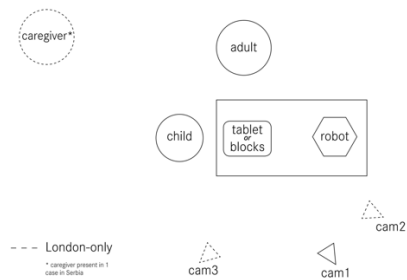

Figure 1: Study setup.

We sourced and modified tangibles that are safe for children ages 0+. The certified blocks used here, from Rubbabu.com, are completely natural and certified for ages 0+. Over 10 potential options for child-safe tangibles were explored, including: child-safe paint, child-safe markers, expandable foam and over six different manufacturers of child blocks. Almost all other solutions, such as paints and most "child-safe" toys, were not suitable for children below the age of 3. We adhered printed natural cotton to display images of the robot.

*Procedure.* At the start of each session the robot was covered with a blanket. When a child entered, the adult would first show them which activity they were going to do using pictographic descriptions. During the first session the children would participate in a pretest to measure their prior knowledge of facial features and facial expressions. Then the adult revealed the robot, which would then introduce itself and the activity. Children then progressed through various steps of game activities as long as time allowed and they remained engaged. The activities included: robot demonstrating facial expressions modeled on realistic expressions based on the Denver corpus from [12], children identifying robot's facial expressions, children asked to make facial expressions and the robot mirroring the child's facial expressions using facial tracking software from [13], children asked to touch the tablet to prompt the robot to explain a facial feature or facial expression, and children asked to play with tangibles of Zeno or of colorful abstract triangles. In subsequent sessions, children would first repeat the previous step, after which they progressed through the remaining steps. Some children participated in NEPSY-II [14] and DANVA-2 [15] tests during post-study follow-up sessions.

In Belgrade only, children were accompanied to the session by their parents or caregivers, but were not present in the room during the study, except for one child who was particularly anxious. The adult also used a paper-based visual schedule to explain the activities to the child and keep track of what they had done so far and how many activities were left. Additional exploratory activities were also tested in Belgrade, including a mix of tablet content, paper, and wooden puzzle activities.

## III. RESULTS

### A. Realistic vs. caricatured facial expressions

*Corpus-based realistic facial expressions.* Participants' responses to the robot's exaggerated facial expressions of emotion recorded in videos were analysed. Two participants asked for caricatured faces multiple times (i.e., immediately upon seeing the robot and later on during the session), which they saw in a prior test. Comparing the current study with videos of a prior study in which caricatured faces were used, children responded more positively with caricatured faces than the current study's Denver corpus faces: they were visibly more comfortable with the robot as observed by the robot holding their engagement while the child exhibited positive emotion compared to engagement without positive emotion or lack of engagement.

> Recommendation: Use facial expressions that are exaggerated rather than the realistic Denver emotions to interest children – at least initially.

*Live mirroring of the child's facial expressions.* The usability test in London ($N = 8$) evaluated facial expressions of the robot that were realistic by mirroring the child's own face. Two participants were excluded due to technical difficulties in the algorithm not being able to account for hair bangs covering the face and when the child was not sufficiently close to the video camera. One child was excluded as he was preoccupied with the robot's material. The mirroring behaviour of the robot was potentially effective in two out of the remaining five cases. These two participants (UK66, UK55) made unprompted facial expressions toward the robot. One child (UK55) showed a lot of positive affect and played for a lengthy amount of time (Figure 2 Left). The other child (UK66) played with the mirroring for a bit, then asked the robot to stop. The remaining three participants did not produce enough facial expressions for them to learn from observation. Instead of fostering attention to faces, this appears to be a difficult activity, risking disengagement (Figure 2 Right).

> Recommendation: Mirroring children's facial expressions through the robot's face is relatively ineffective largely because children did not understand the mirroring and did not make facial expressions when prompted.

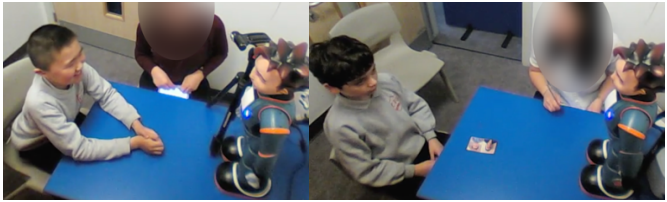
Figure 2. Child responses to mirroring.

There were further issues with the mirroring algorithm. The algorithm attempted to copy the expression, as well as the orientation, of the person's head. As a result, the robot's motion was jerky as it tried to copy the child's motions, which were also short and sudden. The jerkiness made some children anxious. One participant of the three participants who did not produce enough facial expressions for the activity to work (UK54) found the robot's jerky and random motions to be frightening. The jerky motion of the robot may make some children anxious.

Recommendation: Do not include jerky motions in a robot, for example, by smoothing motions.

### B. Tablet and Tangible Peripherals Accompanying Robot

*General use of tablet.* We trialled the use of a tablet, through which children could communicate with the robot and have more input and control over the interaction. Images of the robot's facial features (e.g., eyes) were shown with thick green borders to signify they could select a feature for the robot to demonstrate (Figure 3 Left). Seven out of eight children used the tablet correctly with varying degrees of difficulty (touch being too light or misplaced due to imprecise motor control). This caused frustration or disengagement in some children, who needed to press several times for the system to recognise it. One child appeared afraid of the robot and did not touch the tablet, perhaps because he was aware it would make the robot move.

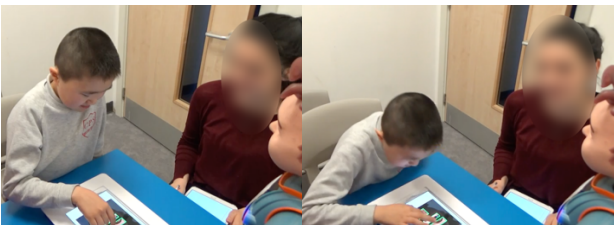
Figure 3. Child responses to tablet.

*Lack of feedback and blocking of tablet interface.* No feedback was provided after selection of a facial feature (e.g., eyes) on the tablet. Although the tablet immediately registered the selection, the latency of the robot's action (e.g., moving and speaking about its eyes) after the child selected took up to three seconds. The time during which the robot processes and performs the action, the tablet UI does not accept input and is unreactive. Four children (UK54,55,60,63) were observed pressing the tablet repeatedly after they had already selected and when the tablet UI was not reactive (Figure 3 Right). Therefore, tablet touch interfaces without press-button feedback and fast latency are ineffective. It causes children to lose interest and may also have impacted their learning of cause and effect (i.e., "my action makes the robot do something"), since the robot did not appear to be consistently responding to their requests.

Recommendation: Implement feedback immediately after a selection is made.

Recommendation: Use a cue to signal to the child when the tablet is inactive (for example, gray out the buttons when they are inactive).

Recommendation: Reduce the time between the robot's response and the child making a selection on the tablet to be very short or negligible (e.g., 1 second or less).

*General use of tangible.* We used a tangible interface in the usability study to gain a better understanding of how the children might engage with the squishy material. Using these materials children engaged in several exploratory game-like activities. They used squishy blocks with the robot's individual facial features to construct a full face, during which the robot would provide encouragement and instructions (Figure 4 Left). Afterwards, they used Lego-like coloured squishy construction blocks to do a free-play tower-building activity together with the robot, during which the robot would prompt the child to find certain colours and stack them together (Figure 4 Right).

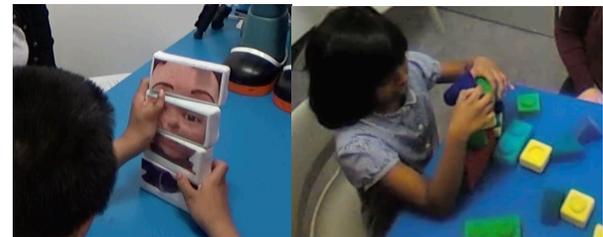
Figure 4: Child responses to tangible "squishies" peripherals used alongside the robot.

We found that children responded very well to the puzzles and paper-based materials. They appeared more focused, and to be engaging more spontaneously. The blocks invite free play and most children immediately start playing with it, including those that do not show much engagement toward the activities that preceded the blocks (i.e., who were not engaged by the robot). The squishy facial feature blocks were often intuitively combined into a full face, although some needed extra help to place individual features in the correct position. Some children labelled the individual features spontaneously or upon request. The coloured blocks were often used to build constructions/towers; however, even children who did not demonstrate the required fine motor skills to stack the Lego-like blocks appeared engaged. Moreover, the material (covered cotton or natural rubber) does not seem to influence this for these children much, as they are using the blocks functionally, rather than focusing primarily on exploring the materials.

Recommendation: Use tangibles to engage children in functional play; cautious use is recommended as the reasons for engagement with the tangibles are unclear.

*Re-engagement with robot while using tangibles.* We also investigated whether children were able to re-engage with the Zeno robot when Zeno spoke to them while they were playing with the blocks. Children who were not engaged by the robot prior to their playing with the blocks, but who were engaged with the blocks, ignored the prompts from Zeno and focussed on solitary play instead (UK63,69). Four out of six children who were engaged by the robot prior to their playing with the blocks were able to redirect their attention back to the robot (UK55,59,60,66) (Figure 5). Therefore, re-engagement

Identify applicable funding agency here. If none, delete this text box.

appeared to be dependent on the child's attention to the robot rather than the attention to the blocks. This means that children who have not demonstrated an inclination to ignore the robot will likely be able to disengage from the materials, but sometimes need some prompting (from adult or robot).

Recommendation: Use of tangibles does not hinder children's ability to re-engage with a robot when the robot speaks.

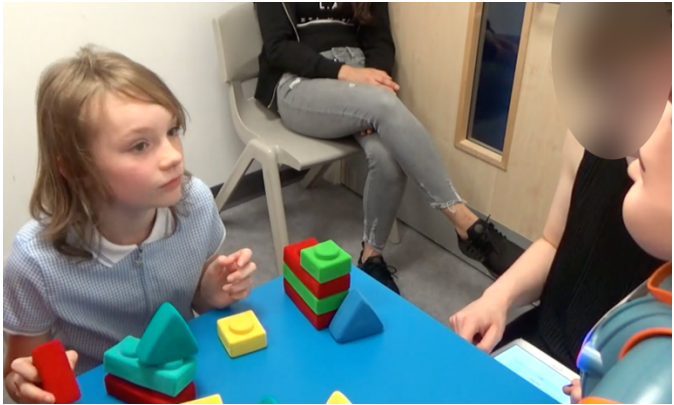

Figure 5: Child re-engagement (UK60) after the robot speaks.

## IV. Discussion

The present research suggests that realistic facial expressions made by a robot are not effective for autistic children who are participating in an emotion learning activity and who need to be engaged. Realistic facial expressions – the types of faces that people actually make when they are expressing emotions, rather than caricatured versions of emotional faces – are important for autistic children to recognize. It is possible that realistic facial expressions could be more useful as advanced activities for children who have already been engaged in the activities.

Tangibles seem promising as children appeared to engage spontaneously with them. However, the reason for this is unclear. One reason could be the quick pacing of the activities or the immediate feedback provided by the activities. For example, during the task in which the child was asked to assemble the robot's face using blocks, the children were given several quick, simple instructions, followed by immediate positive feedback from the adult when they moved and stacked the blocks and a smooth transition to the next task. This is a much quicker sequence than when the robot is used to provide instructions and feedback. In this case, the adult pauses to initiate the robot, the child switches attention from the robot to the tablet and responds, the robot then respond to the child with positive feedback, and in turn, moves on to the next task. The interaction with the tangibles is therefore much more interesting and motivating for the child, with fewer opportunities for distraction or disengagement. If pacing is the major reason for the increased engagement, then we need to improve the pacing with the robot and tablet.

Another reason could be the tangibles themselves. The engagement could be explained by having more tangible items that the child can stack and touch in the adult-led parts of the Belgrade study and a clearer or more interesting task to complete. In this case, it would be important to understand why tangible types of interaction appear more successful than other forms of interaction, in order to improve our design. One possibility is to explore alternative games on the tablet that give the child a sense of achievement and make clear to them that they are progressing toward completing a task. Another possibility is a smart-mat that incorporates both tangible "squishies" and a tablet surface.

## Acknowledgment

This project is funded by the European Union's Horizon 2020 research and innovation programme, grant no. 688835 (DE-ENIGMA). We thank Lynn Packwood for assistance making the tangible block coverings.